\documentclass[runningheads]{llncs}
\usepackage{subcaption}
\usepackage{graphicx}
\usepackage[export]{adjustbox}
\usepackage[T1]{fontenc}
\usepackage{graphicx}
\usepackage{hyperref}
\usepackage{color}

\usepackage{times}
\usepackage{latexsym}
\usepackage{lineno}

\usepackage[utf8]{inputenc}

\usepackage{microtype}
\usepackage{inconsolata}

\newcommand{\framework}{MASGCN}
\usepackage{multicol}
\usepackage{amsfonts}
\usepackage{amsmath}
\usepackage{multirow}
\usepackage{array}
\usepackage{booktabs}
\usepackage{tabularx}
\usepackage{rotating}
\usepackage{enumitem}
\usepackage{amssymb}

\begin{document}
\title{Multi-View Attention Syntactic Enhanced Graph Convolutional Network for Aspect-based Sentiment Analysis}
\titlerunning{MASGCN for Aspect-based Sentiment Analysis}
% If the paper title is too long for the running head, you can set
% an abbreviated paper title here
%
% \author{First Author\inst{1}\orcidID{0000-1111-2222-3333} \and
% Second Author\inst{2,3}\orcidID{1111-2222-3333-4444} \and
% Third Author\inst{3}\orcidID{2222--3333-4444-5555}}
% %
% \authorrunning{F. Author et al.}
% % First names are abbreviated in the running head.
% % If there are more than two authors, 'et al.' is used.
% %
% \institute{Princeton University, Princeton NJ 08544, USA \and
% Springer Heidelberg, Tiergartenstr. 17, 69121 Heidelberg, Germany
% \email{lncs@springer.com}\\
% \url{http://www.springer.com/gp/computer-science/lncs} \and
% ABC Institute, Rupert-Karls-University Heidelberg, Heidelberg, Germany\\
% \email{\{abc,lncs\}@uni-heidelberg.de}}
\author{Xiang Huang\inst{1} \and 
Hao Peng\inst{1,5,6,7,8}\thanks{Corresponding author.} \and
Shuo Sun\inst{1} \and
Zhifeng Hao\inst{2} \and
Hui Lin\inst{3} \and
Shuhai Wang\inst{4}}

\institute{School of Cyber Science and Technology, Beihang University
\email{\{huang.xiang,penghao,sun.shuo\}@buaa.edu.cn}\and
University of Shantou
\email{zfhao@stu.edu.cn}\and
China Academic of Electronics and Information Technology
\email{linhui@cetc.com.cn}\and
Shijiazhuang Tiedao Univerisity
\email{wsh36302@126.com}\and
State Key Laboratory of Public Big Data, Guizhou University\and
Guangxi Key Lab of Multi-source Information Mining \& Security, Guangxi Normal University\and
Yunnan Key Laboratory of Artificial Intelligence, Kunming University of Science and Technology,\and
Hangzhou Innovation Institute of Beihang University
}
\authorrunning{Xiang Huang, et al.}

\maketitle
\begin{abstract}
Aspect-based Sentiment Analysis (ABSA) is the task aimed at predicting the sentiment polarity of aspect words within sentences. 
Recently, incorporating graph neural networks (GNNs) to capture additional syntactic structure information in the dependency tree derived from syntactic dependency parsing has been proven to be an effective paradigm for boosting ABSA.
Despite GNNs enhancing model capability by fusing more types of information, most works only utilize a single topology view of the dependency tree or simply conflate different perspectives of information without distinction, which limits the model performance.
To address these challenges, in this paper, we propose a new multi-view attention syntactic enhanced graph convolutional network (\framework{}) that weighs different syntactic information of views using attention mechanisms. 
Specifically, we first construct distance mask matrices from the dependency tree to obtain multiple subgraph views for GNNs. 
To aggregate features from different views, we propose a multi-view attention mechanism to calculate the attention weights of views. 
Furthermore, to incorporate more syntactic information, we fuse the dependency type information matrix into the adjacency matrices and present a structural entropy loss to learn the dependency type adjacency matrix.
Comprehensive experiments on four benchmark datasets demonstrate that our model outperforms state-of-the-art methods.
The codes and datasets are available at \url{https://github.com/SELGroup/MASGCN}.
\end{abstract}

\section{Introduction}\label{sec:intro}

Aspect-based sentiment analysis (ABSA) is a fine-grained task aimed at classifying the sentiment polarity of given aspect words in a sentence~\cite{absa_survey_tkde}. 
It has drawn significant attention in recent years~\cite{absa_sigir,sigir2023}.
Unlike traditional sentence-level sentiment analysis, ABSA deals with sentences that may contain several different aspect terms in one sentence, each with different sentiment polarities and contexts.
For example, in the sentence \textit{Our waiter was friendly and it is a shame that he didn't have a supportive staff}, the aspect term \textit{waiter} is positive, while \textit{staff} is negative in sentiment.
Consequently, compared with traditional sentiment analysis, ABSA faces the challenge of understanding the relevant context words for different aspects and fully utilizing the related information. 

Early works on ABSA utilized various attention mechanisms~\cite{ataelstm_emnlp,IAN_ijcai2017p568,RAM_emnlp,MGAN_emnlp,huang2018,gu-etal-2018-position,li-etal-2018-hierarchical} to model the relation between the aspect term and its context words. 
Although they have achieved good results, the attention-based weights also introduce substantial noise for aspect terms.
For example, in the sentence \textit{Our waiter was friendly and it is a shame that he didn't have a supportive staff}, for the aspect \textit{waiter}, both opinion words \textit{friendly} and \textit{supportive} may be assigned large attention scores~\cite{rdgcn_wsdm}, which hampers performance improvement.

The rapid development of syntactic parsing offers a new paradigm for ABSA tasks. 
Recently, numerous works~\cite{ASGCN_emnlpijcnlp,tdgat_emnlpijcnlp,bigcn_emnlp,kumagcn_emnlp,dgedt_acl,rgat_acl,dualgcn_aclijcnlp,zhang-etal-2022-ssegcn,LIANG2022107643,rdgcn_wsdm,peng2024prompt} leverage graph neural networks (GNNs) over the parsed dependency tree to exploit the syntactic structure, as the dependency tree provides explicit connections between aspect words and their related opinion words.
These models could be divided into three classes based on distinct syntactic information within the adjacency matrix~\cite{rdgcn_wsdm}: exploiting the topology of trees~\cite{ASGCN_emnlpijcnlp,tdgat_emnlpijcnlp,dgedt_acl}, exploiting the dependency types in trees~\cite{bigcn_emnlp,rgat_acl}, and exploiting the minimum distances in trees~\cite{zhang-etal-2022-ssegcn,rdgcn_wsdm}.
However, they still do not fully exploit the syntactic information of different types. 
~\cite{ASGCN_emnlpijcnlp} and~\cite{tdgat_emnlpijcnlp} only exploit the simple topology information of dependency trees.
\cite{rdgcn_wsdm} consider distance and type information via deep reinforced learning, but they generate only one view for distance information, limiting its expression capability.
\cite{zhang-etal-2022-ssegcn} take multiple views for distance information into account but fuse different views indiscriminately. 
This simple corporation method limits the model to capture as much view information, as it falls short in filtering the introduced noise from more views~\cite{zhang-etal-2022-ssegcn}.
Hence, fully utilizing the diverse syntactic information in dependency trees remains challenging.

Recently, the theory of structural entropy~\cite{li2016structural} has demonstrated significant advantages in graph structural learning~\cite{zou2023se,liu2019rem}. 
It converts the graph into a hierarchical tree, named the encoding tree, by minimizing the graph's structural entropy, which is the sum of the structural entropy of each node in the encoding tree. 
The structural entropy is calculated based on the adjacency matrix and the hierarchical partitions of the graph, where the weights assigned to the matrix are determined by the similarities of the graph node embeddings.
Each type of graph partition has its optimal encoding tree, corresponding to the minimal structural entropy. 
Thus, structural entropy could serve as a graph structural learning target~\cite{user}.
In the ABSA task, syntactic parsing offers an inherent dependency type partition within the dependency tree, which could be leveraged to learn the dependency type information matrix using the structural entropy theory.

In this paper, we propose a novel multi-view attention syntactic enhanced graph convolutional network (\framework{}) for ABSA, which effectively incorporates information from various syntactic views. 
We construct distance mask matrices from the dependency tree to obtain multiple distance information views.
To fuse these distinct views, we propose a multi-view attention mechanism that assigns different attention weights to various views, enhancing the important views and reducing noise from less relevant ones. 
Additionally, to incorporate the syntactic information of dependency types, we introduce a structural entropy loss derived from~\cite{li2016structural}, exploiting the dependency type in the dependency tree to learn the dependency type matrix.
Extensive experiments are conducted on four benchmark datasets.
Comparative results and analysis demonstrate that the proposed model enjoys superior effectiveness compared to the state-of-the-art (SOTA) baselines.

In summary, the main contributions of our paper are as follows:
\begin{itemize}
    \item[$\bullet$] We propose a novel multi-view attention mechanism syntactic enhanced graph convolutional network, which fully leverages extensive syntactic information and mitigates the noise introduced by multiple views.
    \item[$\bullet$] We present a multi-view attention mechanism to weigh different views, reinforce strongly correlated views, and diminish noise from weakly correlated views.
    Additionally, we introduce a structural entropy-based loss to learn the dependency type matrix by utilizing the dependency type information inherent in the dependency tree.
    \item[$\bullet$] We conduct extensive experiments on four benchmark datasets. The experimental results demonstrate the effectiveness of our model.
\end{itemize}

\section{Related Works}
\subsection{Aspect-based Sentiment Analysis}
In contrast to traditional sentiment tasks that are sentence-level or document-level, ABSA is entity-level oriented and more fine-grained for sentiment polarity analysis.
Early works~\cite{jiang-etal-2011-target,kiritchenko-etal-2014-nrc,ding2015deep} extract sentiment features based on handcrafted rules and perform poorly in capturing rich sentiment information.

% attention methods, context information between aspect term and its context
Recently, various context-based methods~\cite{ataelstm_emnlp,IAN_ijcai2017p568,RAM_emnlp,MGAN_emnlp,huang2018,gu-etal-2018-position,li-etal-2018-hierarchical,tan-etal-2019-recognizing} propose to utilize attention mechanisms to model the contextual semantic information between the aspect term and its context words.
ATAE-LSTM~\cite{ataelstm_emnlp} proposes an attention-based LSTM model to concentrate on different parts of sentences to generate attention vectors for aspect sentiment classification.
IAN~\cite{IAN_ijcai2017p568} learns attention between the contexts and targets in an interactive manner and generates representations for targets and contexts separately.
RAM~\cite{RAM_emnlp} leverages the multiple-attention mechanism to capture sentiment features separated by a long distance.
MGAN~\cite{MGAN_emnlp} proposes a fine-grained attention mechanism to capture the word-level interaction between aspects and context and then compose it with coarse-grained attention mechanisms.
\cite{li-etal-2018-hierarchical} design a hierarchical attention mechanism to fuse the information of the aspect terms and the contextual words.
\cite{tan-etal-2019-recognizing} propose a dual attention mechanism to address the problem of recognizing conflicting opinions.
Despite these context-based methods achieving good results, they are unable to distinguish the relation between the aspect term and multiple opinion words.
This hampers their performance in sentences having multiple aspects with different polarities.

% GNN methods, as multiple aspect terms problem.
Owing to the rapid development of syntactic parsing methods, another trend explicitly utilizes the parsed dependency tree to reveal the connection between aspect terms and opinion words and learn the syntactic features of aspect terms.
These methods could be categorized into three classes based on the syntactic information extracted from the dependency tree: exploiting topology information~\cite{ASGCN_emnlpijcnlp,tdgat_emnlpijcnlp,dgedt_acl}, exploiting dependency type information~\cite{bigcn_emnlp,rgat_acl}, and exploiting minimum distance information~\cite{zhang-etal-2022-ssegcn,rdgcn_wsdm}.
\cite{ASGCN_emnlpijcnlp} is the first work to utilize the dependency tree of sentences and adopt GCNs to explore the topological information from the dependency trees.
\cite{dgedt_acl} jointly consider the flat representations learned from the Transformer and the graph-based representations learned from the corresponding dependency graph to diminish the noise induced by incorrect dependency trees.
\cite{bigcn_emnlp} utilize a global lexical graph to encode corpus-level word co-occurrence data and construct a concept hierarchy on both syntactic and lexical graphs, thereby incorporating dependency type information.
\cite{LIANG2022107643} integrate the external knowledge from SenticNet to enhance the dependency graphs of sentences.
\cite{zhang-etal-2022-ssegcn} construct the syntactic mask matrices of different distances to learn structural distance information from local to global.
\cite{wu2023improving} incorporates domain knowledge, dependency labels, and syntax path in the dependency tree to enhance the accuracy of the model.
\cite{rdgcn_wsdm} employ deep reinforcement learning to guide the extraction of syntactic information from the dependency tree.
However, as these approaches only incorporate part of these three syntactic information types, or indiscriminately fuse them, they don't fully leverage the diverse syntactic information in dependency trees.

\subsection{Structural Entropy}
Information entropy was proposed to meet the demand for measuring uncertainty in information transmitted through communication systems.
Correspondingly, to measure the information uncertainty in graph-structured data, structural entropy was also proposed and used to evaluate the complexity of the hierarchical structure of a graph by defining the encoding tree and structural entropy~\cite{li2016structural}. 
The process of constructing and optimizing the encoding tree is also a natural vertices clustering method for graphs.
Due to the theoretical completeness and interpretability of structural entropy theory, it has great potential for application in graph analyses such as graph hierarchical pooling~\cite{wu2022structural} and graph structure learning~\cite{zou2023se,duan2024structural}.
Moreover, the two-dimensional and three-dimensional structural entropy, which measure the complexity of hierarchical structures at two and three dimensions, respectively, have found applications in fields such as medicine~\cite{li2016three}, bioinformatics~\cite{li2018decoding}, social bot detection~\cite{pengunsupervised2024,zeng2024adversarial,yang2024sebot}, network security~\cite{li2016resistance}, natural
language understanding~\cite{huang2024structural} and reinforcement learning ~\cite{zeng2023effective}.
\section{The proposed framework}\label{sec:method}
This section outlines the overall architecture of \framework{} and details each component of our proposed model.

\subsection{Overall Architecture}
\begin{figure*}[t]
\centering
\includegraphics[width=1\textwidth]{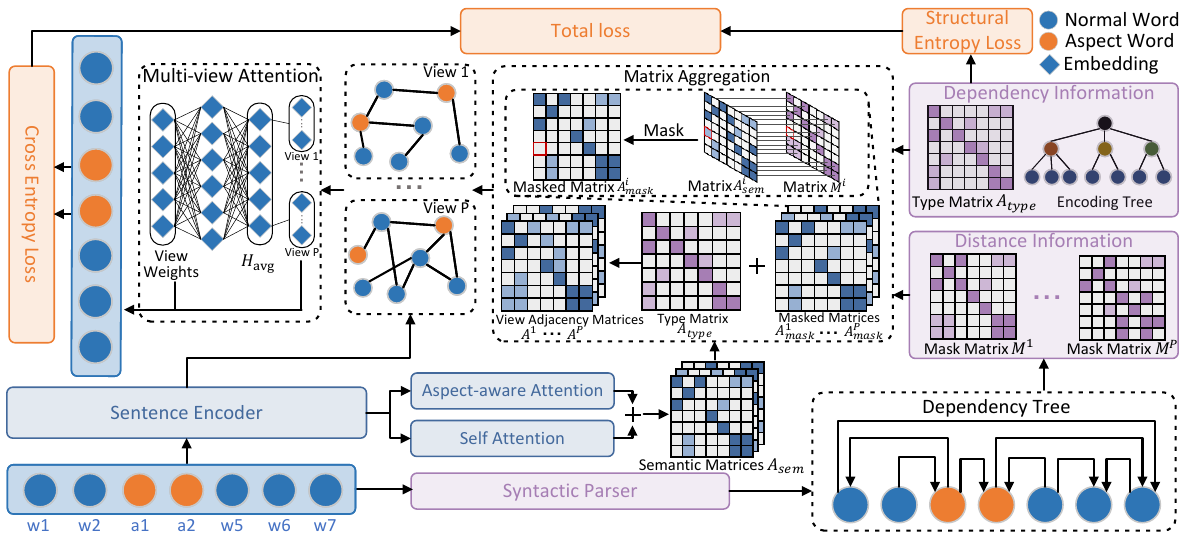}
\caption{Overall framework of \framework{}.}
\label{fig:framework}
\end{figure*}

We illustrate the overall structure of \framework{} in Figure~\ref{fig:framework}.
First, following the approach of~\cite{zhang-etal-2022-ssegcn}, we utilize the aspect-aware attention and self-attention mechanisms to obtain enhanced semantic matrices. 
Second, we parse the sentence syntactically to obtain a dependency tree, constructing distance mask matrices and a dependency type matrix to inform our model of the tree structure.
We also design a structural entropy loss to capitalize on the dependency type information in the dependency tree.
Third, we mask the semantic matrices with distance mask matrices and incorporate the dependency type matrix to create multi-view adjacency matrices for the GNN.
Finally, we propose a multi-view attention mechanism to integrate features across these various views.
% In this section, we introduce the overall architecture of \framework{}. Its overview is shown in Figure\ref{fig:framework}, which includes: extracting semantic features and attention matrices from sentences, utilizing a syntactic parser to obtain the dependency tree, creating distance mask matrices, learning a dependency type matrix along with a structural entropy loss, masking the semantic attention matrix with distance masks and the dependency type matrix to form various view adjacency matrices for the GNN, and employing Multi-view Attention Mechanisms to combine multi-view features.

\subsection{Semantic Feature}\label{subsec:semantic feature}
Given a sentence-aspect pair $\mathcal{W}-\mathcal{A}$, where $\mathcal{A} = \{a_1, \dots, a_M\}$ represents the aspect words set and is also a sub-sequence of the sentence $\mathcal{W} = \{w_1, \dots, w_N\}$, with $w_i$ and $a_j$ denoting the $i$-th and $j$-th words in $\mathcal{W}$ and $\mathcal{A}$, and $N$ and $M$ representing the lengths of the sentence and the aspect, respectively.
We derive the contextual feature of sentences based on the sentence encoder like BERT~\cite{bert} and obtain the low-dimensional embedding matrix $H\in \mathbb{R}^{N\times D}$, where the $i$-th word $w_i$ corresponds to the $i$-th row feature $h_i$ with dimension $D$.
Aspect features $H_a\in \mathbb{R}^{M\times D}$ are derived from $H$.

For enhanced semantic features, we follow~\cite{zhang-etal-2022-ssegcn} and apply the aspect-aware attention and self-attention mechanism to obtain aspect attention matrices $\{A_{\text{asp}}^1, \dots, A_{\text{asp}}^P\}$ and self-attention matrices $\{A_{\text{self}}^1, \dots, A_{\text{self}}^{P}\}$, where $P$ is the number of attention heads.
The $i$-th matrices are calculated as follows:
\begin{align}
    \label{eq:aspect_aware_attention}
    A_{\text{asp}}^i &= \tanh\left(\hat{H}_a W_a^i\times\left(HW_k^i\right)^T + b_a\right),\\
    A_{\text{self}}^i &= \frac{QW_Q^i\times (KW_K^i)^T}{\sqrt{D}}.  
\end{align}
Here, $\hat{H}_a\in\mathbb{R}^{N\times D}$ represents the $N$ times repeated mean of $H_a$, $W_a^i\in \mathbb{R}^{D\times D}$ and $W_k^i\in \mathbb{R}^{D\times D}$ are learnable weights for aspect-aware attention of the $i$-th attention head, and $b_a$ is the learnable bias. 
The query $Q$ and the key $K$ are equal to the feature embedding $H$, and $W_Q^i\in \mathbb{R}^{D\times D}$ and $W_K^i\in \mathbb{R}^{D\times D}$ are learnable weights for the self-attention of the $i$-th attention head.
Then, we integrate $A_{\text{asp}}^i$ and $A_{\text{self}}^i$ as follows:
\begin{equation}
    \label{eq:semantic matrix}
    A_{\text{sem}}^i = A_{\text{asp}}^i + A_{\text{self}}^i,
\end{equation}
where $A_{\text{sem}}^i\in\mathbb{R}^{N\times N}$ is the output enhanced semantic matrix. These $P$ matrices $\{A_{\text{sem}}^1, \dots, A_{\text{sem}}^P\}$ are used for later combination with syntactic matrices.

\subsection{Syntactic Feature}\label{subsec:syntactic feature}

In this section, we first utilize the Stanford Parser CoreNLP~\cite{corenlp} to parse the sentence into a syntactic dependency tree, and then extract the distance information and dependency information from the dependency tree.

\noindent\emph{\textbf{$\bullet$ Distance information.}} 
We treat the syntactic dependency tree as an undirected graph, with each word as a node.
Then, we define the distance $d(v_i, v_j)$ between nodes $v_i$ and $v_j$ as the number of hops between the two nodes in the graph.
The shortest path distance between nodes $v_i$ and $v_j$ is calculated as follows:
\begin{equation}\label{eq:minimal distance}
    D_{ij} = \min d(v_i, v_j).
\end{equation}
We consider different scales of distance information by defining various distance mask matrices $M^k$ as follows:
\begin{equation}\label{eq:distance mask}
    M_{ij}^{k} = \left\{
    \begin{aligned}
        0, \quad&D_{ij} \leq k,\\
        -\infty, \quad&\text{otherwrise,}
    \end{aligned}
    \right.
\end{equation}
where $k\in [1, P]$ and $P$ is the number of attention heads mentioned in Section~\ref{subsec:semantic feature}.
As $k\in [1, P]$ increases, the scope of distance information that the mask matrix $M^k$ covers also increases.

\noindent\emph{\textbf{$\bullet$ Dependency information.}} 
We define the initial dependency type adjacency matrix $A_{\text{type}}^0\in \mathbb{R}^{N\times N}$ as follows:
\begin{equation}\label{eq:initial dependency matrix}
    A_{\text{type}}^0[i, j] = \left\{
    \begin{aligned}
        id_{\text{type}}, \quad &D_{ij} = 1,\\
        0, \quad &\text{otherwise},
    \end{aligned}
    \right.
\end{equation}
where $id_{\text{type}}\in [1, U]$ and $U$ is the size of the dependency type vocabulary.
To utilize the dependency information of the dependency tree, we initialize the dependency type feature matrix $H_{\text{type}}\in\mathbb{R}^{U\times D}$.
The global attention mechanism~\cite{rdgcn_wsdm} calculates the dependency type attention matrix as follows:
\begin{align}\label{eq:dependency global attention mechanism}
    \alpha &= \text{softmax}(H_{\text{type}}W_t),\\
    A_{\text{type}}[i,j] &= \left\{
    \begin{aligned}
        \alpha[A_{\text{type}}^0[i,j]], \quad&A_{\text{type}}^0[i,j]\neq 0,\\
        0, \quad&\text{otherwise},\\
    \end{aligned}
    \right. 
\end{align}
where $W_t\in\mathbb{R}^{D\times 1}$ represents the learnable weights for global attention, $\alpha\in \mathbb{R}^{U\times 1}$ denotes the attention score.
$A_{\text{type}}\in \mathbb{R}^{N\times N}$ is the dependency type information matrix output.

\noindent\emph{\textbf{$\bullet$ Matrix Aggregation.}}
To this end, we have obtained the semantic feature matrices $\{A_{\text{sem}}^1, \dots, A_{\text{sem}}^P\}$, the distance information mask matrices $\{M^1, \dots, M^P\}$, and the dependency type information matrix $A_{\text{type}}$. 
We then aggregate these semantic and syntactic information to construct the $i$-th view adjacency matrices for GNN as follows:
\begin{align}\label{eq:adjacency matrix}
    A^i &= A^i_{\text{mask}} + A_{\text{type}},\\ A^i_{\text{mask}} &= \text{softmax}(A_{\text{sem}}^i+M^i).
\end{align}
Here, we utilize the softmax operator to mask the $i$-th semantic feature matrix $A_{\text{sem}}^i$ with the $i$-th distance mask matrix $M^i$, as $-\infty$ in $M^i$ masks the corresponding place in $A_{\text{sem}}^i$ to $0$.
As $i$ increases, $A^i_{\text{mask}}$ could attend more global information.
Finally, we obtain $P$ distinct view adjacency matrices $\{A^1, \dots, A^P\}$, each attending to different semantic and syntactic information.

\subsection{GNN and Multi-View Attention Mechanism}
Now we have obtained the sentence embedding $H$ and $P$ distinct adjacency matrices $\{A^1, \dots, A^P\}$. 
The output embedding of the $l$-th GNN layer is calculated as follows:
\begin{equation}\label{eq:gnn}
    H_l = \sigma\left(AGG_{i=1}^{P}(A^i H_{l-1})W_l\right), H_0 = H,
\end{equation}
where $H_l\in\mathbb{R}^{N\times D}$ is the $l$-th layer output embedding, $W_l\in\mathbb{R}^{D\times D}$ represents the $l$-th layer’s learnable weights, and $\sigma$ denotes an activation function. 
The operator $AGG_{i=1}^{P}$ is the multi-view attention mechanism we propose, which is as follows:
\begin{align}\label{eq:multi-view attention}
    &H_{\text{avg}}^i = \frac{1}{N\times D}\sum_{r, s=1}^{N ,D}\left(A^iH_{l-1}\right)_{r, s},\\
    &\alpha_{\text{view}} = W_2\left(\sigma\left(W_1H_{\text{avg}}\right)\right),\\
    &AGG_{i=1}^P(A^iH_{l-1}) = \sum_{i=1}^{P}\alpha_{\text{view}}^iA^iH_{l-1}.
\end{align}
Here, $H^i_{\text{avg}}\in \mathbb{R}$ is $i$-th row of $H_{\text{avg}}\in \mathbb{R}^{P\times 1}$, $\alpha_{\text{view}}\in \mathbb{R}^{P\times 1}$ is the view weights of $P$ views, $W_1$ and $W_2$ are the learnable weight, $\alpha^i_{\text{view}}$ is the $i$-th row of $\alpha_{\text{view}}$.
The multi-view attention mechanism adaptively extracts useful information from various views, and leverages the diversity and complementarity of information across different perspectives, thereby enhancing the representation capabilities for downstream tasks.
Additionally, the multi-view attention mechanism reduces noise introduced by too many views~\cite{zhang-etal-2022-ssegcn}.

\subsection{Loss Function with Structural Entropy}
% To further enhance the utilization of syntactic information in the dependency tree, we design a structural entropy loss.
Structural entropy theory~\cite{li2016structural} has demonstrated its capability in multiple graph-related works~\cite{wu2022structural,zou2023se,duan2024structural}.
For a graph $G=(X, E, W)$, $X$ is the set of graph datapoints, and $E$ and $W$ are the edges and corresponding edge weights.
We define a two-level encoding tree $\mathcal{T}$, where the intermediate tree nodes $\alpha$ represent a partitioned subset of the graph vertices $X$, denoted as $X_{\alpha} \subset X$.
The leaf nodes are the graph vertices, and the top layer of $\mathcal{T}$ is a single virtual node.
The original two-dimensional structural entropy of $\mathcal{T}$ is defined as follows:
\begin{align}
    H^{\mathcal{T}}(G) &= \sum_{\alpha \in \mathcal{T}} H^{\mathcal{T}}(G; \alpha),\\
    H^{\mathcal{T}}(G; \alpha) &= -\frac{g_{\alpha}}{\mathrm{vol}(G)}\log_2\frac{\mathcal{V}_{\alpha}}{\mathrm{vol}(G)}.  
\end{align}
Here, $\mathrm{vol}(G)$ is the sum of the edge weights of all edges $E$, $\mathcal{V}_{\alpha}$ represents the sum of the edge weights of edges in $X_{\alpha}$, and $g_{\alpha}$ is the sum of the edge weights of cut edges between $X_{\alpha}$ and its complement set $X_{\alpha}^{\complement}$.

Within the ABSA task, syntactic parsing provides a natural segmentation of dependency types within the dependency tree.
This partitioning can be utilized to develop a matrix of dependency type information based on the principles of structural entropy theory.
To further enhance the utilization of syntactic information in the dependency tree, we design a structural entropy loss $\mathcal{L}_{SE}$.
In detail, we first build the two-layer encoding tree $\mathcal{T}$ with the dependency type as the immediate layer's tree nodes.
The encoding tree is an abstraction of the inherent dependency type partition $\mathcal{P} = \{\mathcal{P}_1, \dots, \mathcal{P}_{U}\}$ of words, where $U$ is the size of the dependency type vocabulary.
Denoting $Y\in\left\{0, 1\right\}^{N\times U}$ as the one-hot encoding of $\mathcal{P}$, the two-dimensional structural entropy for this encoding tree can be described as follows:
\begin{equation}\label{eq:structural entropy}
    \mathcal{L}_\text{SE} = \text{trace}\left\{\frac{Y^TA_{\text{type}}Y}{2\sum A_{\text{type}}}\otimes \log_{2}\frac{\{1\}^{U\times N}A_{\text{type}}Y}{2\sum A_{\text{type}}}\right\},
\end{equation}
where $\otimes$ refers to matrix multiplication.
As the dependency type partition $\mathcal{P}$ is fixed, the optimized matrix $A_{\text{type}}$ can be obtained as the minimal two-dimensional structural entropy~\cite{user}.
Combined with cross-entropy loss, the final training loss is as follows:
\begin{align}\label{eq:loss}
    \mathcal{L} &= \mathcal{L}_{\text{ce}} + \gamma \mathcal{L}_\text{SE},\\
    \mathcal{L}_{\text{ce}} = -\sum_{a\in\mathcal{D}}\log&\left(\text{softmax}\left(W_pH_{a}^l + b_p\right)\right),
\end{align}
where $W_p$ and $b_p$ are learnable weights and bias, $H_a^l$ is the aspect embeddings of the final $l$-th GNN layer, $\mathcal{D}$ is the set of traing samples, and $\gamma$ is a hyperparameter.

\section{Experiment Setup}
In this section, we detail the baselines, datasets, and implementation details of \framework{}.

\subsection{Datasets}

Following previous ABSA works, we evaluate \framework{} on four benchmark datasets: Restaurant14, Restaurant16, Laptop14, and Twitter. 
The Restaurant14 and Laptop14 datasets include reviews in the restaurant and laptop domains from SemEval-2014~\cite{semeval2014-semeval}. 
The Restaurant16 dataset is from SemEval-2016~\cite{semeval2016-semeval}.
The Twitter dataset is collected from tweets by~\cite{tweets-acl}. 
Each aspect is annotated with one of three polarities: positive, neutral, and negative.
The statistics of these datasets are listed in Table~\ref{tab:datasets}.

\begin{table}[t]
    \aboverulesep=0ex
    \belowrulesep=0ex
    \centering
    \setlength{\tabcolsep}{0.8mm}
    \caption{Statistics of experimental datasets.}
    \begin{tabular}{lcccc}
    \toprule
    Dataset & Division & Positive & Negative & Neutral\\
    \hline
    \multirow{2}{*}{Restaurant14} & Train & 2164 & 807 & 637\\
    & Test & 727 & 196 & 196\\
    \hline
    \multirow{2}{*}{Laptop14} & Train & 976 & 851 & 455\\
    & Test & 337 & 128 & 167\\
    \hline
    \multirow{2}{*}{Twitter} & Train & 1507 & 1528 & 3016\\
    & Test & 172 & 169 & 336\\
    \hline
    \multirow{2}{*}{Restaurant16} & Train & 1321 & 488 & 73\\
    & Test & 487 & 136 & 32\\
    \bottomrule
    \end{tabular}
    \label{tab:datasets}
\end{table}

\subsection{Baselines}
To comprehensively evaluate the performance of \framework{}, we compare it with SOTA baselines, which are briefly described as follows:

\noindent 1) \textit{Context-based methods.}

\noindent $\bullet$ \textbf{ATAE-LSTM~\cite{ataelstm_emnlp}} is an LSTM model with the attention mechanism on aspects.

\noindent $\bullet$ \textbf{IAN~\cite{IAN_ijcai2017p568}} interactively learns attention scores for aspects with their context. 

\noindent $\bullet$ \textbf{RAM~\cite{RAM_emnlp}} proposes a recurrent attention memory network to capture the aspect-specific sentence representation.

\noindent $\bullet$ \textbf{MGAN~\cite{MGAN_emnlp}} applies a fine-grained attention mechanism to capture token-level interactions between aspects and contexts.

\noindent $\bullet$ \textbf{TNet~\cite{TNet_acl}} utilizes a CNN model to extract salient features for sentiment analysis.

\noindent 2) \textit{Syntax-based GNN methods.}

\noindent $\bullet$ \textbf{ASGCN~\cite{ASGCN_emnlpijcnlp}} applies GCN on the raw topology of the dependency tree to extract syntactic information.

\noindent $\bullet$ \textbf{kumaGCN~\cite{kumagcn_emnlp}} uses gating mechanisms to acquire syntactic features with latent semantic information.

\noindent $\bullet$ \textbf{DGEDT~\cite{dgedt_acl}} combines transformer and graph-based representations from the corresponding dependency graph to diminish the error induced by incorrect dependency trees.

\noindent $\bullet$ \textbf{BiGCN~\cite{bigcn_emnlp}} uses convolutions on hierarchical lexical and syntactic graphs to integrate token co-occurrence and dependency type information.

\noindent $\bullet$ \textbf{R-GAT~\cite{rgat_acl}} uses a star-induced graph with minimum distances and dependency types as edges and applies a relational GAT for attention-based aggregation.

\noindent $\bullet$ \textbf{T-GCN~\cite{tgcn_naaclhlt}} uses attention to distinguish relation types and applies an attentive layer ensemble for feature learning from GCN layers.

\noindent $\bullet$ \textbf{DualGCN~\cite{dualgcn_aclijcnlp}} introduces syntactic and semantic information through SynGCN and SemGCN modules simultaneously.

\noindent $\bullet$ \textbf{SSEGCN~\cite{zhang-etal-2022-ssegcn}} combines aspect-aware attention, self-attention, and minimum tree distances to enhance sentiment information with syntactic information.

\noindent $\bullet$ \textbf{SenticGCN~\cite{LIANG2022107643}} integrates external knowledge from SenticNet to enhance the dependency graphs of sentences.

\noindent $\bullet$ \textbf{SPGCN~\cite{lu2022sentiment}} enhances dependencies and context-awareness with aspect focus and sentiment integration, using a multi-graph perception mechanism to reduce redundancy and captures unique dependencies.

\noindent $\bullet$ \textbf{KDGN~\cite{wu2023improving}} integrates domain knowledge into the dependency tree and constructs the knowledge-aware dependency graph to capture latent sentiment polarity.

\noindent $\bullet$ \textbf{CSADGCN~\cite{CSADGCN}} integrates context-guided, self, and aspect-level attention for information capture and boosts GCN with linguistic features and a biaffine module for word relationships.

\noindent $\bullet$ \textbf{RDGCN~\cite{rdgcn_wsdm}} utilizes deep reinforced learning to extract syntactic information and improve the importance calculation of dependencies in both distance and type views.

\noindent $\bullet$ \textbf{BERT \& Model+BERT~\cite{bert}} represents the pre-trained language model BERT and the model with BERT as the sentence encoder.

\subsection{Experimental Settings}
In experiments, we initialize token embeddings with pre-trained 300-dimensional Glove vectors~\cite{glove-EMNLP}, combine them with 30-dimensional part-of-speech and position embeddings, and feed them into a BiLSTM model with a 0.7 dropout rate.
The batch size is 16 and the number of GCN layers is 2.
The learning rate of the Adam optimizer is 0.002.
The view number and attention head number $P$ are set to 10.
The hyperparameter $\gamma$ of the structural entropy loss is 0.01.
Model+BERT utilizes the bert-base-uncased~\cite{bert} English version.
We utilize accuracy (Acc.) and macro-F1 (F1) to evaluate classification performance.
The baseline results are from~\cite{rdgcn_wsdm} and~\cite{kga_tkde}.

\begin{table*}[t]
    \aboverulesep=0ex
    \belowrulesep=0ex
    \centering
    \caption{Classification results (\%) of all models. The best results are bolded, and the second-best results are underlined. $\dagger$ denotes topology information, $\ddagger$ denotes type information, and $\S$ denotes distance information.}
    \setlength{\tabcolsep}{1mm}
    \begin{tabular}{clcccccccc}
         \toprule
         \multirow{2}{*}{Embedding} & \multirow{2}{*}{Model} & \multicolumn{2}{c}{Restaurant14} & \multicolumn{2}{c}{Laptop14} & \multicolumn{2}{c}{Twitter} & \multicolumn{2}{c}{Restaurant16} \\
         \cline{3-10} & & Acc. & F1  & Acc. & F1 & Acc. & F1 & Acc. & F1\\
         
         \hline
         \multirow{13}{0.065\textwidth}[-2ex]{\centering Glove} &ATAE-LSTM & 78.60  & 67.02 & 68.88  & 63.93 & 68.64 & 66.60 & 83.77 & 61.71\\
         &IAN & 78.60   & - & 72.10   & - & - & - & - & -\\
         &RAM & 80.23 & 70.80 & 74.49 & 71.35 & 69.36 & 67.30 & 83.88 & 62.14\\
         &MGAN & 81.25 & 71.94 & 75.39 & 72.47 & 72.54 & 70.81 & - & - \\
         &TNet & 80.69 & 71.27 & 76.54 & 71.75 & 74.90 & 73.60 & 89.07 & 70.43\\
         \cline{2-10}
         % &PWCN\textsuperscript{$\S$} & 80.96 & 72.21 & 76.12 & 72.12 & - & -\\
         & ASGCN\textsuperscript{$\dagger$} & 80.77 & 72.02 & 75.55 & 71.05 & 72.15 & 70.40 & 88.69 & 66.64\\
         % &TD-GAT\textsuperscript{$\dagger$} & 81.20 & - & 74.00 & - & - & -\\
         &kumaGCN\textsuperscript{$\dagger$} & 81.43 & 73.64 & 76.12 & 72.42 & 72.45 & 70.77 & 89.39 & 73.19\\ 
         &DGEDT\textsuperscript{$\dagger$} & 83.90 & 75.10 & 76.80 & 72.30 & 74.80 & 73.40 & \underline{90.80} & 73.80\\
         & BiGCN\textsuperscript{$\dagger\ddagger$}  & 81.97 & 73.48 & 74.59 & 71.84 & 74.16 & 73.35 & - & -\\
         &R-GAT\textsuperscript{$\dagger\ddagger\S$} & 83.30 & 76.08 & 77.42 & 73.76 & 75.57 & 73.82 & 88.92 & 70.89\\
         &DualGCN\textsuperscript{$\dagger$} & 84.27 & \underline{78.08} & 78.48 & 74.74 & 75.92 & 74.29 & - & -\\ 
         & SSEGCN\textsuperscript{$\dagger\S$} & \underline{84.72} & 77.51 & 79.43 & 76.49 & 76.51 & 75.32 & 89.55 & 75.62\\
         & SenticGCN\textsuperscript{$\dagger\ddagger$} & 84.03 & 75.38 & 77.90 & 74.71 & - & - & \textbf{90.88} & \underline{75.91}\\
         & SPGCN\textsuperscript{$\dagger\ddagger\S$} & 83.16  & 74.91 & 77.90 & 73.86 & 74.86 & 72.95 & 90.75 & 75.20\\
         &RDGCN\textsuperscript{$\dagger\ddagger\S$} & 84.36 & 78.06 & \underline{79.59} & \underline{76.75} & \underline{76.66} & \underline{75.37} & - & -\\
         % \framework{} & Topology \& Type \& Distance & \textbf{85.25} & \textbf{78.87} & \textbf{79.75} & 76.02 & \textbf{77.25} & \textbf{75.98}\\
         &\textbf{\framework{}}\textsuperscript{$\dagger\ddagger\S$} & \textbf{84.99} & \textbf{78.64} & \textbf{80.38} & \textbf{77.74} & \textbf{76.96} & \textbf{75.85} & 89.01 & \textbf{77.13}\\

         \hline

        \multirow{8}{0.065\textwidth}[-2ex]{\centering \shortstack{BERT}}&BERT & 85.97 & 80.09 & 79.91 & 76.00 & 75.92 & 75.18 & 89.52 & 70.47\\
        &DGEDT+BERT\textsuperscript{$\dagger$} & 86.30 & 80.00 & 79.80 & 75.60 & 77.90 & 75.40 & 91.90 & 79.00\\
        &R-GAT+BERT\textsuperscript{$\dagger\ddagger\S$} & 86.60 & 81.35 & 78.21 & 74.07 & 76.15 & 74.88 & 89.71 & 76.62\\
        &T-GCN+BERT\textsuperscript{$\dagger\ddagger$} & 86.16 & 79.95 & 80.88 & 77.03 & 76.45 & 75.25 & \underline{92.32} & 77.29\\
        &DualGCN+BERT\textsuperscript{$\dagger$} & 87.13 & 81.16 & 81.80 & 78.10 & 77.40 & 76.02 & - & - \\
        &SSEGCN+BERT\textsuperscript{$\dagger\S$} & 87.31 & 81.09 & 81.01 & 77.96 & 77.40 & 76.02 & 90.99 & 78.78\\
        &SenticGCN+BERT\textsuperscript{$\dagger\ddagger$} & 86.92 & 81.03 & 82.12 & 79.05 & - & - & 91.97 & \underline{79.56}\\
        & KDGN+BERT\textsuperscript{$\dagger\ddagger$} & 87.01 & \underline{81.94} & 81.32 & 77.59 & 77.64 & 75.55 &-&-\\
        & CSADGCN+BERT\textsuperscript{$\dagger\ddagger\S$} & 87.40 & 81.56 & 82.12 & \textbf{79.22} & 76.81 & 75.67 & - & -\\
        &RDGCN+BERT\textsuperscript{$\dagger\ddagger\S$} & \underline{87.49} & 81.16 & \underline{82.12} & 78.34 & \underline{78.29} & \textbf{77.14} & - & -\\ 
        % \framework{}+BERT & Topology \& Type \& Distance & \textbf{87.58} & \textbf{82.52} & \textbf{82.12} & \textbf{78.41} & \textbf{78.58}& \textbf{77.34}\\
        &\textbf{\framework{}+BERT}\textsuperscript{$\dagger\ddagger\S$} & \textbf{87.76} & \textbf{82.56} & \textbf{82.44} & \underline{79.11} & \textbf{78.58} & \underline{77.05} & \textbf{92.52} & \textbf{80.72}\\

         \bottomrule
    \end{tabular}
    \label{tab:performance table}
\end{table*}
\section{Experiments}\label{sec:experiments}
% In this section, we first introduce the datasets and experimental settings. 
In this section, we conduct comparative experiments with baselines for effectiveness comparison. 
Moreover, we carry out the ablation study, parameter sensitivity experiments, and investigations into adjacency matrices to explore each component of our model.
% This section covers dataset introduction, experimental setup, baseline comparisons, ablation study, and parameter sensitivity analysis for our model.

\subsection{Performance Comparision}

\begin{table*}[ht]
    \aboverulesep=0ex
    \belowrulesep=0ex
    \centering
    \caption{Ablation study results (\%). The best results are bolded, and the second-best results are underlined. w/o refers to the model without the corresponding component.}
    \setlength{\tabcolsep}{1mm}
    \begin{tabular}{lcccccccc}
         \toprule
         \multirow{2}{*}{Model} & \multicolumn{2}{c}{Restaurant14} & \multicolumn{2}{c}{Laptop14} & \multicolumn{2}{c}{Twitter} & \multicolumn{2}{c}{Restaurant16} \\
         \cline{2-9} & Acc. & F1  & Acc. & F1 & Acc. & F1 & Acc. & F1\\
         \hline
         \framework{} & \textbf{84.99} & \textbf{78.64} & \textbf{80.38} & \textbf{77.74} & \underline{76.96} & \textbf{75.85} & \textbf{89.01} & \textbf{77.13}\\
         - w/o structural entropy loss $\mathcal{L}_{SE}$ & \underline{84.63} & \underline{78.07} & \underline{79.73} & \underline{76.88} & \textbf{77.25} & \underline{75.61} & \underline{89.01} & \underline{76.92}\\
         - w/o multi-view attention mechanism & 83.56 & 75.98 & 79.11 & 75.01 & 75.63 & 74.43 & 88.23 & 69.44\\
         \bottomrule
    \end{tabular}
    \label{tab:case study}
\end{table*}
We conduct the experiments on four benchmark datasets and report the experiment results in Table~\ref{tab:performance table}. 
\framework{} achieves state-of-the-art results on all four datasets, except for accuracy on the Restaurant16 dataset when using Glove embeddings.
Specifically, with Glove embeddings, \framework{} outperforms all baselines with improvements of at least 0.17–0.79 in accuracy and 0.48–0.99 in macro-F1 across the Restaurant14, Laptop14, and Twitter datasets.
On the Restaurant16 dataset, \framework{} still excels in macro-F1, showing a 1.22 improvement over the SOTA SenticGCN model.
Besides, \framework{} + BERT surpasses other Model+BERT combinations across all datasets, except for macro-F1 on the Twitter dataset and the Laptop14 dataset.
These results demonstrate the effectiveness of our model. 
% Moreover, as the output of the BERT model contains rich semantics, the performance gain of \framework{} + BERT over baselines is less than \framework{} alone.
% In general, models that utilize topology, type, and distance information perform better.
Compared to R-GAT, SPGCN, CSADGCN, and RDGCN, which also utilize three types of information, \framework{} still outperforms due to its enhanced utilization of this information through multi-view information aggregation.

\begin{figure*}[t]
    \centering
    \includegraphics[width=1\textwidth]{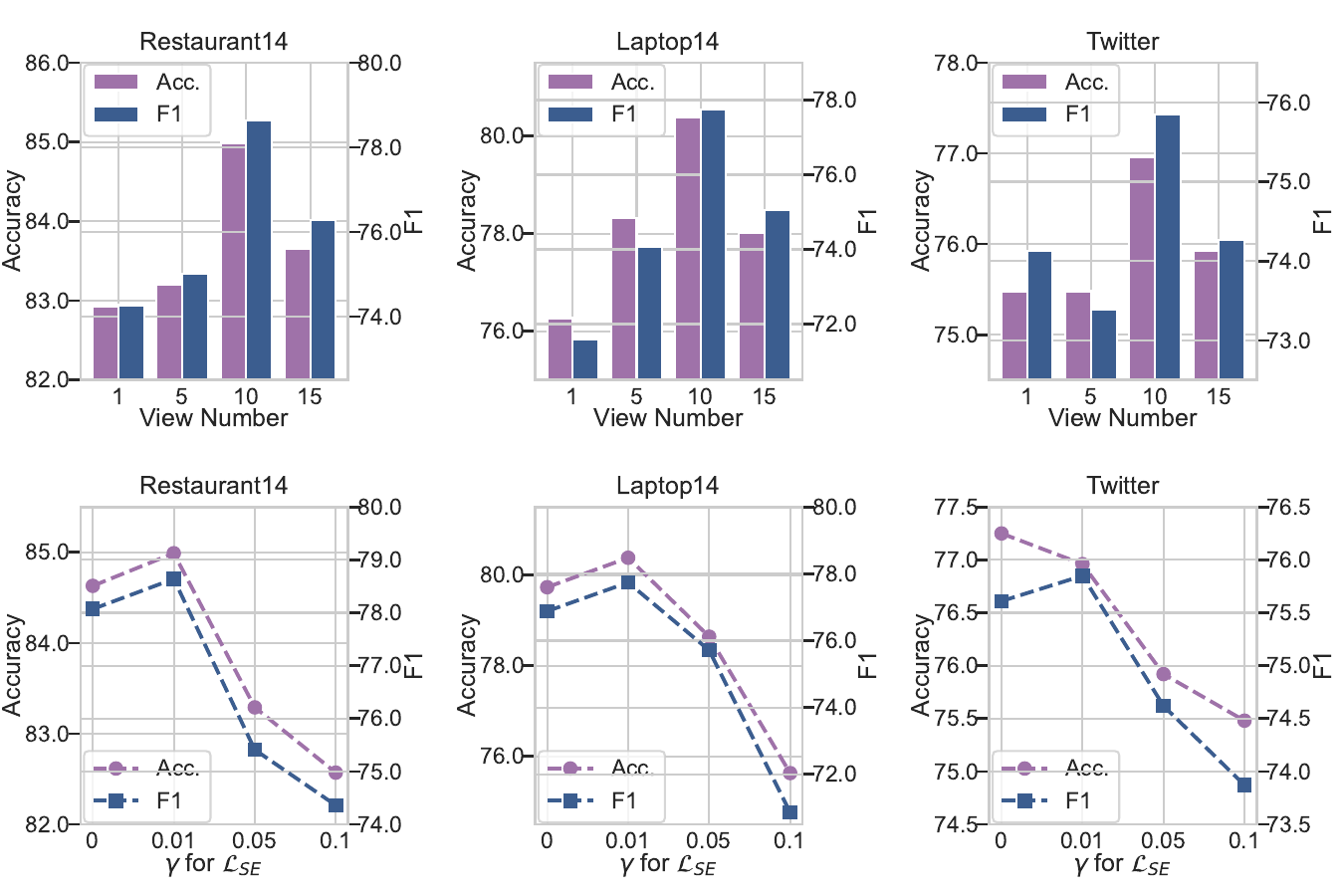}
    \caption{Hyperparameter sensitivity analysis.}
    \label{fig:hyperparameter}
\end{figure*}

\begin{figure*}[t]
    \centering
    \begin{subfigure}{\linewidth}{
        \includegraphics[width=1\linewidth]{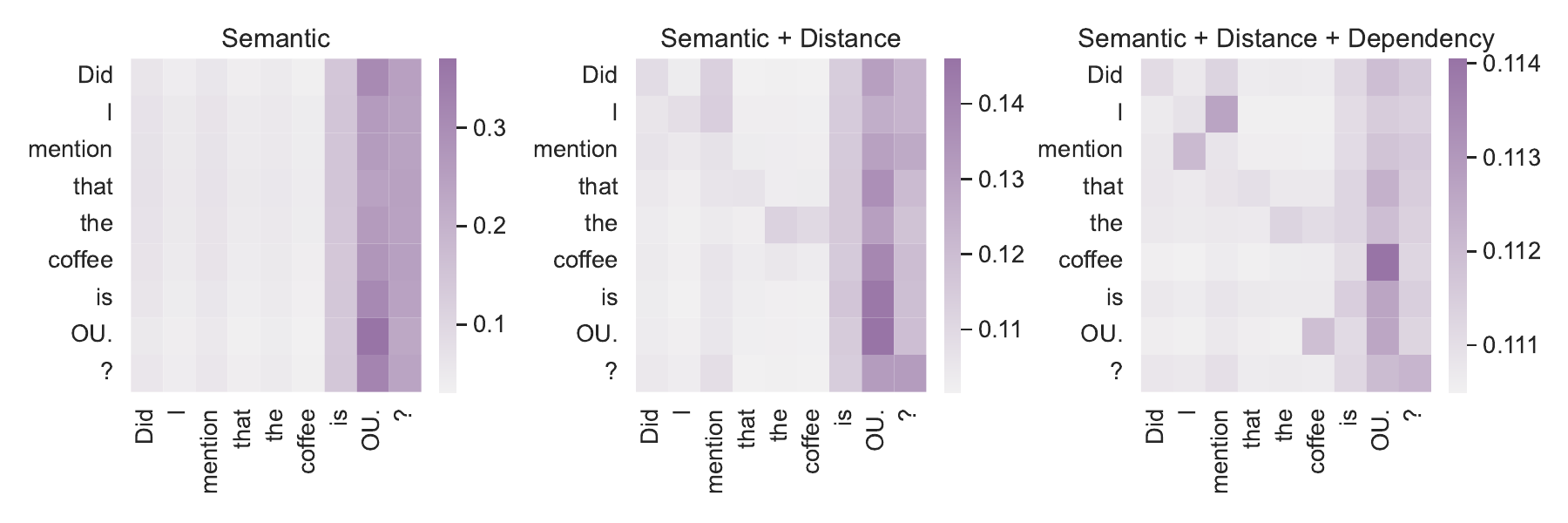}
        \caption{The aspect term is \textit{coffee}. }
    }
    \end{subfigure}
    \begin{subfigure}{\linewidth}{
        \includegraphics[width=1\linewidth]{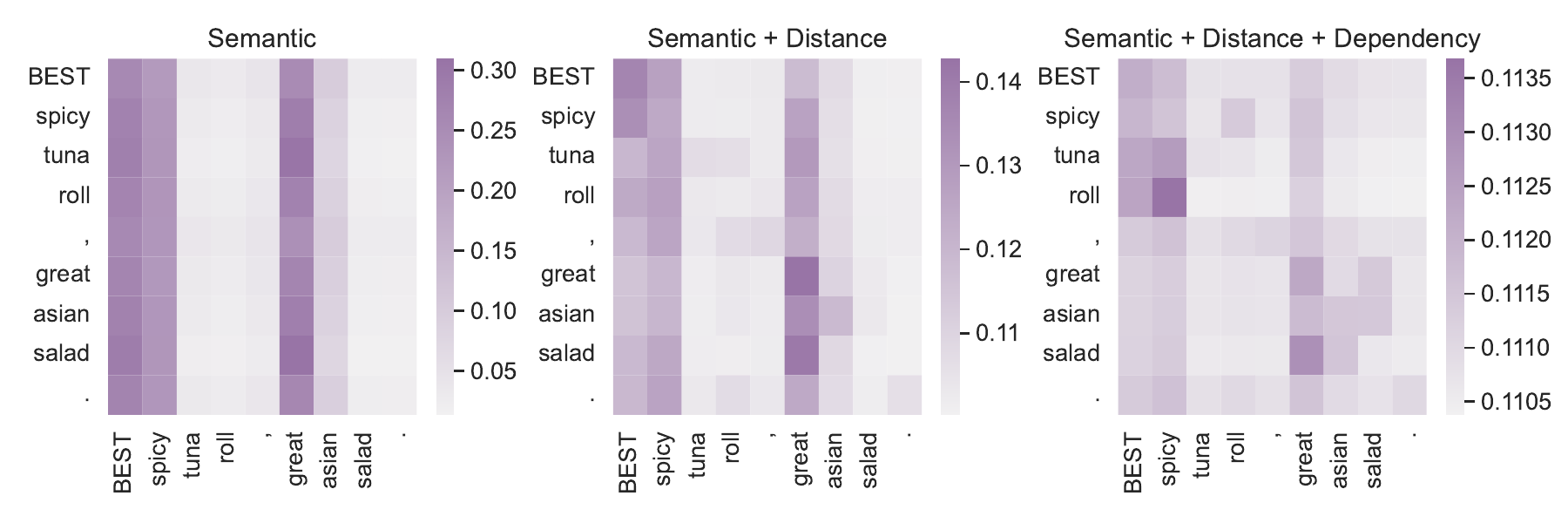}
        \caption{The aspect terms are \textit{roll} and \textit{salad}.}
    }
    \end{subfigure}
    \caption{Insights of matrices.}
    \label{fig:score matrices}
\end{figure*}

\subsection{Ablation Study}
We evaluate each component of \framework{} on four datasets and report the results in Table~\ref{tab:case study}. 
The absence of structural entropy loss $\mathcal{L}_{SE}$ and the multi-view attention mechanism results in a decline in model performance on most datasets, with an average reduction of 0.18\% and 1.20\% in accuracy, and 0.47\% and 3.63\% in macro-F1, respectively.
This demonstrates the effectiveness of structural entropy loss in learning dependency type information and our proposed multi-view attention mechanism in fusing syntactic information.

\subsection{Hyperparameter Analysis}
Two hyperparameters are introduced in our model: the view number $P$ and the coefficient $\gamma$ for the structural entropy loss $\mathcal{L}_{SE}$.
We vary their values on three benchmark datasets and illustrate the results in Figure~\ref{fig:hyperparameter}.
When the view number $P$ is not greater than 10, more views improve the model performance.
However, the performance decreases when $P$ is 15. 
This is because the number of views exceeds the maximum distance in the syntax tree, thereby introducing only additional noise.
Despite this, compared to SSEGCN~\cite{zhang-etal-2022-ssegcn}, which fuses different views of distance information indiscriminately and only integrates four different views when balancing the introduction of additional noise, \framework{} still improves the ability to fuse more information and reduce noise.
Furthermore, we find that a smaller $\gamma$ for structural entropy loss is preferred for the model.
\framework{} achieves the best performance with $\gamma$ being 0.01, except on Twitter, where $\gamma$ is 0.
A $\gamma$ larger than 0.01 leads to a decrease in model performance.

\subsection{Insight of Matrices}
To elucidate the effectiveness of each type of information, we visualize the semantic matrix $A_{\text{sem}}$, the masked matrix $A_{\text{mask}}$ (incorporating semantic and distance information), and the final fused adjacency matrix $A$ (incorporating semantic, distance, and dependency information) mentioned in Section~\ref{sec:method}.
As depicted in Figure~\ref{fig:score matrices} (a), with the integration of distance information and dependency information, the correlation of the aspect term \textit{coffee} with the polarity word \textit{OUTSTANDING} is enhanced, and the attention to other uncorrelated words is gradually reduced.
Figure~\ref{fig:score matrices} (b) shows that the integration of distance information and dependency information decreases the irrelevant attentions, enhances the attention of the aspect term \textit{roll} to its polarity words \textit{BEST spicy}, and reduces the attention to the polarity word \textit{great}, which is connected with the other aspect term \textit{salad}.
The improvement can also be observed in the row of the aspect term \textit{salad}.
These two cases directly demonstrate the effectiveness of \framework{}.
\section{Conclusion}\label{sec:conclution}
In this paper, we propose a new multi-view attention syntactic enhanced graph
convolutional network \framework{}. 
We propose a multi-view attention mechanism to aggregate features from different views, which enhances the strongly correlated views while diminishing the noise introduced by too many views.
Additionally, we design a structural entropy based loss to further leverage the dependency type information from the dependency tree.
Extensive experiments on four benchmark datasets demonstrate that \framework{} outperforms the state-of-the-art baseline methods. 

\section{Limitations and Future Work}
Our work incorporates various syntactic information from the dependency tree to improve the effectiveness of ABSA tasks.
However, it is limited to the information from the parsed dependency tree and does not consider external knowledge.
In future work, it would be valuable to exploit information from other models, such as external knowledge graphs or large language models like GPT-4~\cite{openai2024gpt4technicalreport}.
\section*{Acknowledgments}
This work is supported by the NSFC through grants 62322202, 62441612, and 62432006, Local Science and Technology Development Fund of Hebei Province Guided by the Central Government of China through grant 246Z0102G, the "Pioneer” and “Leading Goose” R\&D Program of Zhejiang" through grant 2025C02044, the Hebei Natural Science Foundation through grant F2024210008, and the Guangdong Basic and Applied Basic Research Foundation through grant 2023B1515120020, Foundation of State Key Laboratory of Public Big Data through grant PBD2022-04, and Research Fund of Guangxi Key Lab of Multi-source Information Mining \& Security (MIMS23-M-01).

% Bibliography entries for the entire Anthology, followed by custom entries
%\bibliography{anthology,custom}
% Custom bibliography entries only
\bibliographystyle{splncs04}
\bibliography{custom}

% \appendix

% \section{Example Appendix}
% \label{sec:appendix}

% This is an appendix.

\end{document}